\def\year{2020}\relax
\documentclass[letterpaper]{article} 
\usepackage{aaai20}  
\usepackage{times}  
\usepackage{helvet} 
\usepackage{courier}  
\usepackage[hyphens]{url}  
\usepackage{graphicx} 
\usepackage{graphicx}  
\usepackage{amsmath}
\usepackage{bm}
\usepackage{amsfonts,amssymb}
\newcommand{\norm}[1]{\left\lVert#1\right\rVert}
\usepackage{array}
\usepackage{booktabs}
\usepackage{multirow}

\title{Fast and Robust Face-to-Parameter Translation for Game Character Auto-Creation}
\author{
Tianyang Shi,\textsuperscript{\rm 1} 
Zhengxia Zou,\textsuperscript{\rm 2} 
Yi Yuan,\textsuperscript{\rm 1}\thanks{Corresponding author} 
Changjie Fan\textsuperscript{\rm 1}\\ 
\textsuperscript{\rm 1}NetEase Fuxi AI Lab \\
\textsuperscript{\rm 2}University of Michigan, Ann Arbor\\ 
\{shitianyang, yuanyi, fanchangjie\}@corp.netease.com, zzhengxi@umich.edu
}

\begin{document}

\maketitle

\begin{abstract}
With the rapid development of Role-Playing Games (RPGs), players are now allowed to edit the facial appearance of their in-game characters with their preferences rather than using default templates. This paper proposes a game character auto-creation framework that generates in-game characters according to a player's input face photo. Different from the previous methods that are designed based on neural style transfer or monocular 3D face reconstruction, we re-formulate the character auto-creation process in a different point of view: by predicting a large set of physically meaningful facial parameters under a self-supervised learning paradigm. Instead of updating facial parameters iteratively at the input end of the renderer as suggested by previous methods, which are time-consuming, we introduce a facial parameter translator so that the creation can be done efficiently through a single forward propagation from the face embeddings to parameters, with a considerable 1000x computational speedup. Despite its high efficiency, the interactivity is preserved in our method where users are allowed to optionally fine-tune the facial parameters on our creation according to their needs. Our approach also shows better robustness than previous methods, especially for those photos with head-pose variance. Comparison results and ablation analysis on seven public face verification datasets suggest the effectiveness of our method.\par
\end{abstract}

\newcolumntype{L}[1]{>{\raggedright\arraybackslash}p{#1}}
\newcolumntype{C}[1]{>{\centering\arraybackslash}p{#1}}
\newcolumntype{R}[1]{>{\raggedleft\arraybackslash}p{#1}}

\section{Introduction}

\begin{figure}[ht]
	\centering
	\includegraphics[width=0.95\linewidth]{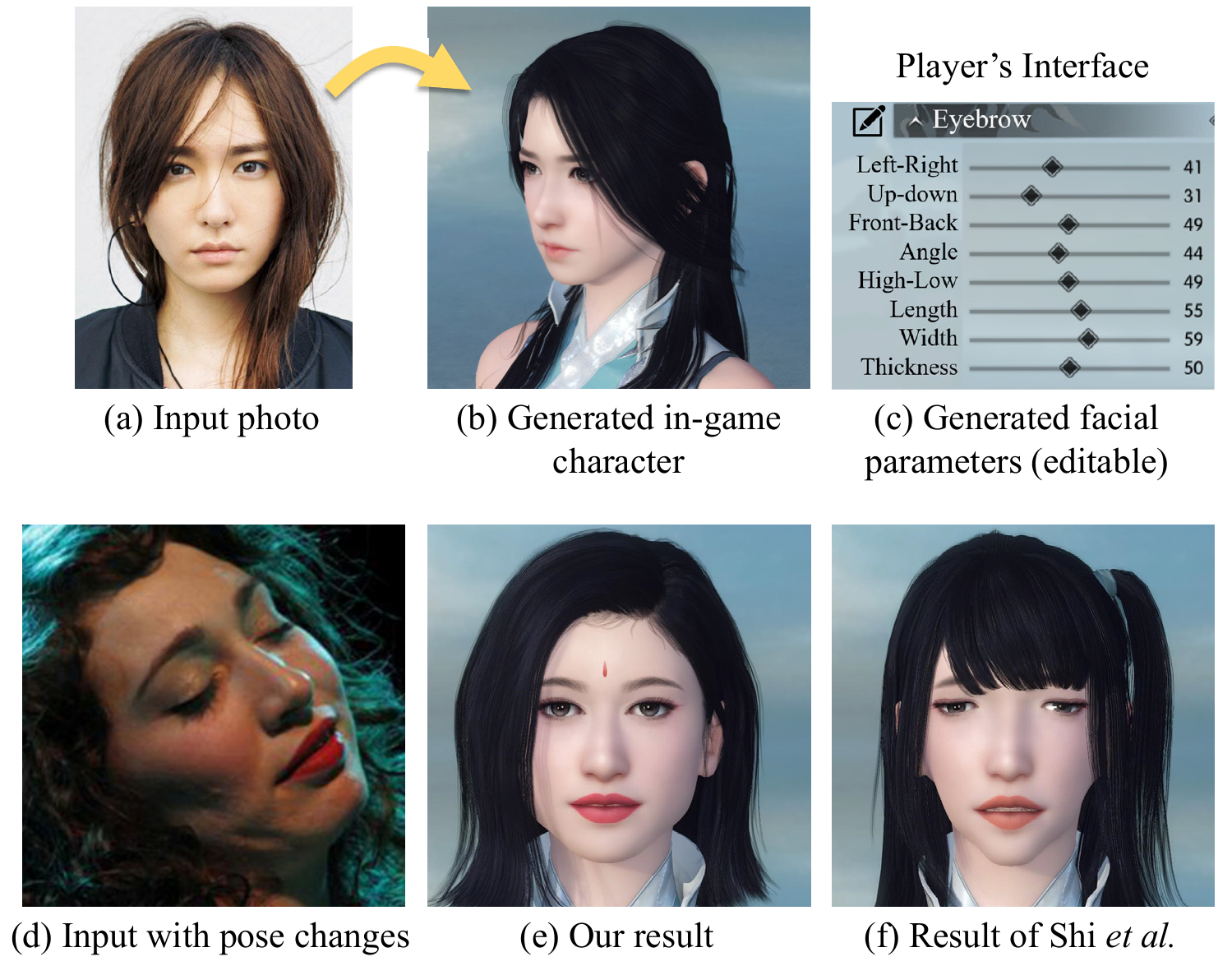}
	\caption{We propose a new method to generate in-game character of the players according to their input face photo. Our method is three magnitudes faster than the previous method \cite{Shi_2019_ICCV} and shows better robustness (especially for pose changes).}
	\label{fig:1stpage_preview}
\end{figure}
\par

The character customization system in many role-playing games (RPGs) provides an interactive interface where players can create their in-game characters according to their preferences. To improve players' immersion and interactivity, character customization systems in RPGs are becoming more and more sophisticated. In many modern RPGs, e.g., ``Grand Theft Auto Online\footnote{https://www.rockstargames.com/GTAOnline}'', ``Dark Souls III\footnote{https://www.darksouls.jp}'', and ``Justice\footnote{https://n.163.com}'', players are now allowed to precisely manipulate specific part of their characters, e.g., the corner of the eye, the wing of the nose, hairstyles, and makeups. As a result, the character customization process turns out to be laborious and time-consuming. To create an in-game character with a desired facial appearance (e.g.\ a pop star or the players themselves), most players need to spend several hours by manually adjusting hundreds of parameters, even after considerable practice. In this paper, we study an interesting problem, i.e. how to automatically generate in-game characters according to a player's face photo.\par

In computer vision, efforts have been made in generating 3D faces based on single input face photo, where some representative methods include the 3DMM \cite{blanz1999morphable} and 3DMM-CNN \cite{Tran_2017_CVPR}, etc. However, these methods are difficult to be applied in-game environments. This is because most of the above methods are built based on ``morphable face models'' \cite{blanz1999morphable} while the faces in most RPGs are rendered by ``bone-driven face models''. There is a huge gap between the two different type of face models due to their different rendering mode, skeletons, and physical meaning of facial parameters. \par

To solve the character auto-creation problem in RPGs, Shi \textit{et al} \cite{Shi_2019_ICCV} proposed a method for bone-driven face model called ``Face-to-Parameters'' (F2P). In the F2P method, a CNN model is trained to make game engine differentiable, and the character auto-creation is formulated as a facial similarity measurement and parameter searching problem. The facial parameters are optimized under a neural style transfer framework \cite{Gatys2016Image,Gu_2018_CVPR}, where they iteratively adjust the parameters at the input end of the renderer by using gradient decent. Despite their promising results, the F2P method still has some limitations in both robustness and processing speed. As the local feature they used dominates the facial similarity measurement, their method is sensitive to pose changes. Besides, since the parameter generation is performed based on an iterative updating process, the F2P suffers from a low inference speed.\par

In this paper, we propose a new framework to solve the above problems. Different from the previous methods which are designed based on neural style transfer \cite{Shi_2019_ICCV} or monocular 3D face reconstruction \cite{blanz1999morphable,Richardson_2017_CVPR,Tewari_2017_ICCV}, we re-formulate the character auto-creation process under a \textit{self-supervised learning} paradigm. Instead of updating facial parameters iteratively at the input end of the renderer, as suggested by Shi \textit{et al.} \cite{Shi_2019_ICCV}, we introduce a facial parameter translator so that the optimal facial parameters can be obtained efficiently through a single time forward propagation. This improvement increases the inference speed of the previous method by three orders of magnitude. Despite its high efficiency, the interactivity is also preserved in our method where users are allowed to optionally fine-tune the generated facial parameters according to their needs. To improve the robustness of our framework, especially for pose variance of the input face, instead of directly predicting facial parameters in their original parameter space, we integrated the facial priors by learning their low-dimensional representation and making predictions in their orthogonal subspace. We shed light on how these simple modifications improvement the robustness of our framework. \par

\begin{figure*}[t]
	\centering
	\includegraphics[width=0.9\linewidth]{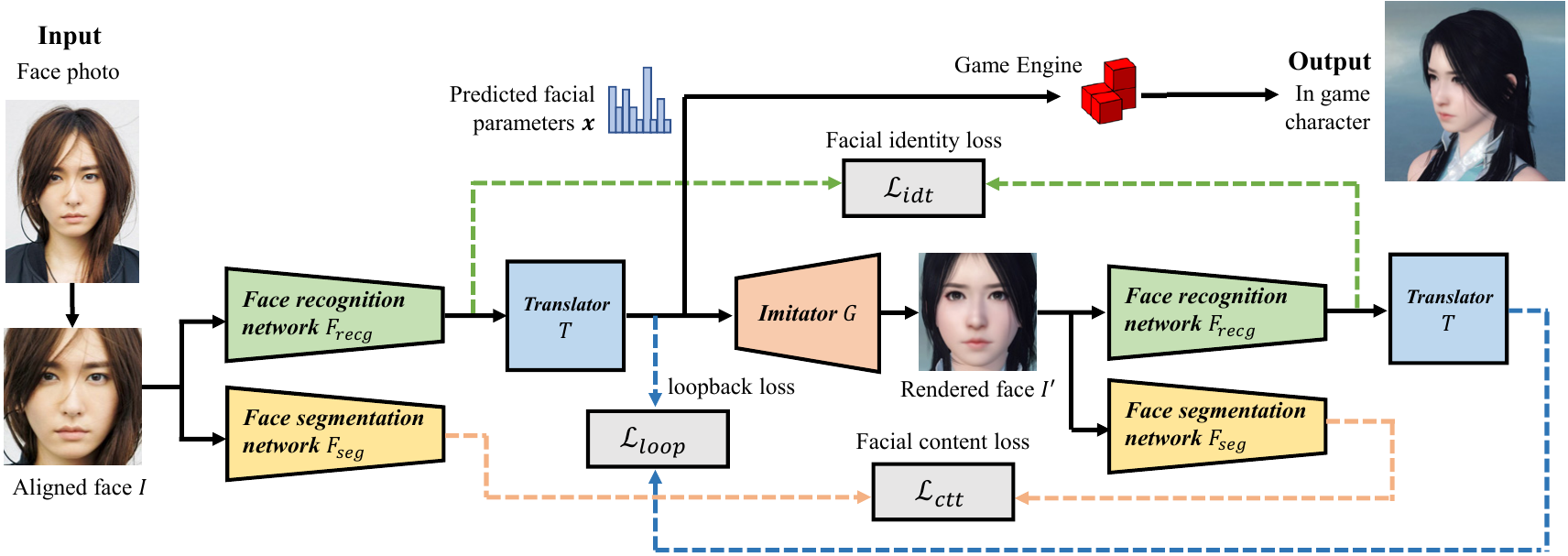}
	\caption{Our model consists of four networks: 1) an Imitator $G$ to imitate the behavior of a game engine, 2) a Translator $T$ to predict facial parameters, 3) a face recognition network $F_{recg}$ to encode an input image to face embeddings, and 4) a face segmentation network $F_{seg}$ to extract position-sensitive facial representations. The key to our method is a self-supervised learning framework where a ``recursive consistency'' is introduced to \textit{enforce} the facial representation of the rendered image $I^\prime$ to be similar with that of its input $I$: $I^\prime =G(T(F_{recg}(I))) \approx I.$}
	\label{fig:pipeline}
\end{figure*}

Fig. \ref{fig:pipeline} shows an overview of the proposed framework. Our framework consists of multiple components:

- \textbf{An imitator $G$}. As the rendering process of most game engines is not differentiable, we train an imitator $G$ to imitate the behavior of a game engine, as suggested by previous methods \cite{Shi_2019_ICCV,genova2018unsupervised}, and make the rendering process differentiable. In this way, our framework can be smoothly trained in an end-to-end fashion.

- \textbf{A translator $T$}, which aims to transform the input facial image embeddings to output facial parameters. The generated parameters either can be further fine-tuned by players or can be used to render 3D faces in-game environments.

- \textbf{A face recognition network $F_{recg}$}, which encodes an input face image to a set of pose-irrelevant face embeddings.

- \textbf{A face segmentation network $F_{seg}$}, which extracts position-sensitive face representations.

We formulate the training of our framework under a multi-task regression paradigm by constructing multiple self-supervised loss functions. To measure the similarity between an input face and a generated one, we define two loss functions, an ``identity loss'' $\mathcal{L}_{idt}$ and a ``facial content loss'' $\mathcal{L}_{ctt}$ for effective facial similarity measurement, where the former one focuses on facial identity (pose-irrelevant) and the later one compute similarity base on pixel-wise representations. To further improve the robustness and stability, we also introduce a ``loopback loss'' $\mathcal{L}_{loop}$, which ensures the translator can correctly interpret its own output. Our contributions are summarized as follows:\par

1) We propose a new framework for game character auto-creation under a self-supervised learning paradigm. The training of our method neither requires any ground truth references nor any user interactions. \par

2) Different from previous method \cite{Shi_2019_ICCV} that the facial parameters are optimized iteratively by gradient descent, we introduce a translator $T$ where the prediction of the facial parameter only requires single time forward-propagation. We improve the inference speed over three orders of magnitude. \par

3) Facial priors are integrated as auxiliary constraints by performing whitening transformation on a large set of facial parameters. An orthogonal space with reduced dimensions is used for improving robustness. \par

\begin{table}[t]
\centering
\caption{A comparison between the Morphable Face Model (MFM) and Bone-Driven Face Model (BDFM).}
\begin{tabular}{p{3cm}p{2cm}p{2cm}}
\toprule
& {\bf MFM}  & {\bf BDFM} \\
\midrule
{\bf Physical Meaning} & ambiguous & explicit \\
{\bf Degree of Freedom} & normal & high \\
{\bf Texture Style} & real & game-style \\
{\bf Structure} & parametrized & bone-driven \\
{\bf Ground Truth} & a few & none \\
{\bf Makeup} & a few & many \\
\bottomrule
\end{tabular}
\label{tab:MFM_vs_BDFM}
\end{table}

\section{Related Work}

\subsection{3D Face Reconstruction} 
3D Face Reconstruction, as a classical ill-posed problem in computer vision, aims to recover 3D information from a 2D photo. A representative of the early method of this field is the well-known 3D Morphable Model (3DMM), which was originally proposed by Blanz \textit{et al.} \cite{blanz1999morphable}. They use principal component analysis (PCA) to parameterize the 3D mesh data and then optimize the parameter of the face model to fit input photo \cite{blanz1999morphable}. On this basis, some improved methods e.g., BFM09 \cite{BFM2009} and BFM17 \cite{BFM2017}, were then proposed to improve the shape, texture, and expression of the morphable model. In recent years, deep Convolutional Neural Networks (CNNs) are also introduced to predict the parameters of the morphable model with the help of high-level image representations \cite{Tran_2017_CVPR,Dou_2017_CVPR,jackson2017large}. Besides, the Generative Adversarial Network (GAN) has also been introduced to this field recently to generate high-fidelity textures \cite{gecer2019ganfit}. \par

As the 3DMM-based reconstruction can be essentially considered as a parameter fitting problem between 3D face model and input facial image, making renderer differentiable is the key to solve this problem. With the development of the differentiable rendering technology, self-supervised loss functions can now be used for training neural network without using pre-scanned 3D faces \cite{Tewari_2017_ICCV,genova2018unsupervised}. \par

But unfortunately, most of the above methods cannot be directly applied to game environments. The reasons are twofold. First, these methods are designed based on morphable face models while in most RPGs, the 3D faces are rendered by the bone-driven face model. Second, the morphable face model is not friendly for user interactions as each of its parameters lacks a clear physical meaning. Table \ref{tab:MFM_vs_BDFM} gives a list of differences between these two models. \par

\subsection{Character Auto-Creation} 
Character auto-creation is of great importance for computer games and augmented reality. As an emerging technology, it has drawn increasing attention in recent years \cite{wolf2017unsupervised,Shi_2019_ICCV}. Wolf \textit{et al.} \cite{wolf2017unsupervised} first proposed an adversarial training based method named Tied Output Synthesis (TOS) for creating parameterized avatars based on face photos, in which they take advantage of adversarial training to select the facial components from a set of pre-defined templates. However, this method is designed to predict discrete attributes rather than continuous facial parameters. A similar work to ours is ``Face-to-Parameter (F2P)'' method \cite{Shi_2019_ICCV} proposed by Shi \textit{et al.}, where they build their method based on the bone-driven face model and frame the character creation as a neural style transfer process. To make the renderer differentiable, they train a generative network to imitate the input-output behavior of a game engine. They also introduce two discriminative networks to measure and maximize the similarity between the input photo and the generated character.\par

\section{Methodology}

Our model consists of four neural networks: a imitator $G$, a facial parameter translator $T$, a pre-trained face recognition network $F_{recg}$, and a face segmentation network $F_{seg}$, as shown in Fig.~\ref{fig:pipeline}.

\subsection{Imitator}

We train a convolutional neural network as our imitator to fit the input-output mapping of a game engine so that to make the character customization system differentiable. We take the similar configuration with the generator of DCGAN \cite{radford2015unsupervised} in our imitator $G(\bm{x})$, which also has been used in F2P \cite{Shi_2019_ICCV}. We frame the training of our imitator as a standard regression problem, where we aim to minimize the difference between the in-game rendered image and the generated one in their pixel space. When training the imitator, we use pixel-wise $l_1$ loss function as it encourages less blurring effect:
\begin{equation}\label{eq:G_loss}
    \mathcal{L}_G (\bm{x}) = E_{\bm{x}\sim u(\bm{x})}\{ {\|G(\bm{x}) - \text{Engine}(\bm{x})\|}_1 \},
\end{equation}
where $\bm{x}$ is the facial parameters sampled from a uniform distribution. Given $\bm{x}$ as an input, the $G(\bm{x})$ and $\text{Engine}(\bm{x})$ are the outputs of our imitator and a game engine (ground truth), respectively. For simplicity, our imitator $G$ only fits the front view of the facial model with the corresponding facial parameters.\par

\begin{figure}[t]
	\centering
	\includegraphics[width=0.9\linewidth]{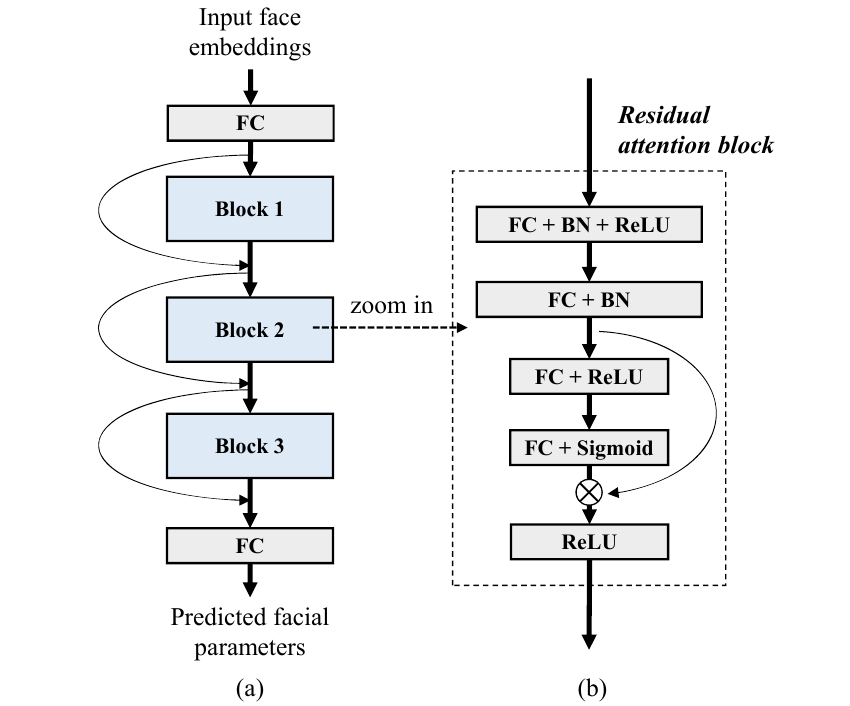}
	\caption{(a) The architecture of our translator $T$. Our translator consists of three residual attention blocks and two fully connected layers (FC). (b) The details of the residual attention block. $\bigotimes$ denotes element-wise product between two neural activations. }
	\label{fig:translator}
\end{figure}

\subsection{Translator}

We train a multi-layer perceptron as our translator $T$ to map the face embeddings $\bm{f}_{recg}$ (the output of our face recognition network) to facial parameters $\bm{x}$. 
\begin{equation}\label{eq:T_inference}
    \bm{x} = T(\bm{f}_{recg}) = T(F_{recg}(I)),
\end{equation}
where $F_{recg}$ is the face recognition network, and $I$ is an input face photo. Since the face embeddings correspond to a global description of the input identity while the facial parameters depict local facial details, to learn better correspondence of the two fields, we introduce an attention-based module as the core building block of our translator. Our translator consists of three residual attention blocks for generating internal representations and two Fully Connected (FC) layers for input and output. The details of our translator are shown in Fig.~\ref{fig:translator}. \par

The attention mechanism was originally proposed in machine translation to improve the performance of an RNN model \cite{bahdanau2014neural}. It has now been widely used in many computer vision tasks, such as object detection \cite{zhang2018occluded}, image super-resolution \cite{zhang2018image}, etc. To introduce the attention mechanism in our translator, we compute the element-wise importance of the neurons and then perform feature re-calibration by multiplying the representations with them. We make a simple modification of the squeeze and excitation block in SENet \cite{hu2018squeeze} to apply it in an FC layer (the global pooling layer thus is removed). Specifically, let $\bm{v}\in \mathbb{R}^{c\times 1}$ represents a $c$ dimensional internal representation of our translator. We use a gating function to learn the element-wise attention weights, which can be written as: 
\begin{equation}
    \bm{\alpha} = \sigma(\bm{W}_2 \delta(\bm{W}_1 \bm{v})),
\end{equation}
where $\sigma(\cdot)$ and $\delta(\cdot)$ denote the sigmoid and ReLU activation function, respectively. $\bm{W}_1$ and $\bm{W}_2$ are the weights of two FC layers. Then, internal representation $\bm{v}$ is re-scaled as follows: 
\begin{equation}
    \widetilde{v}_k = \alpha_k v_k,
\end{equation}
where $\alpha_k$, $k=1,\dots c$ are the learned attention weights. 

\begin{figure}[t]
	\centering
	\includegraphics[width=0.85\linewidth]{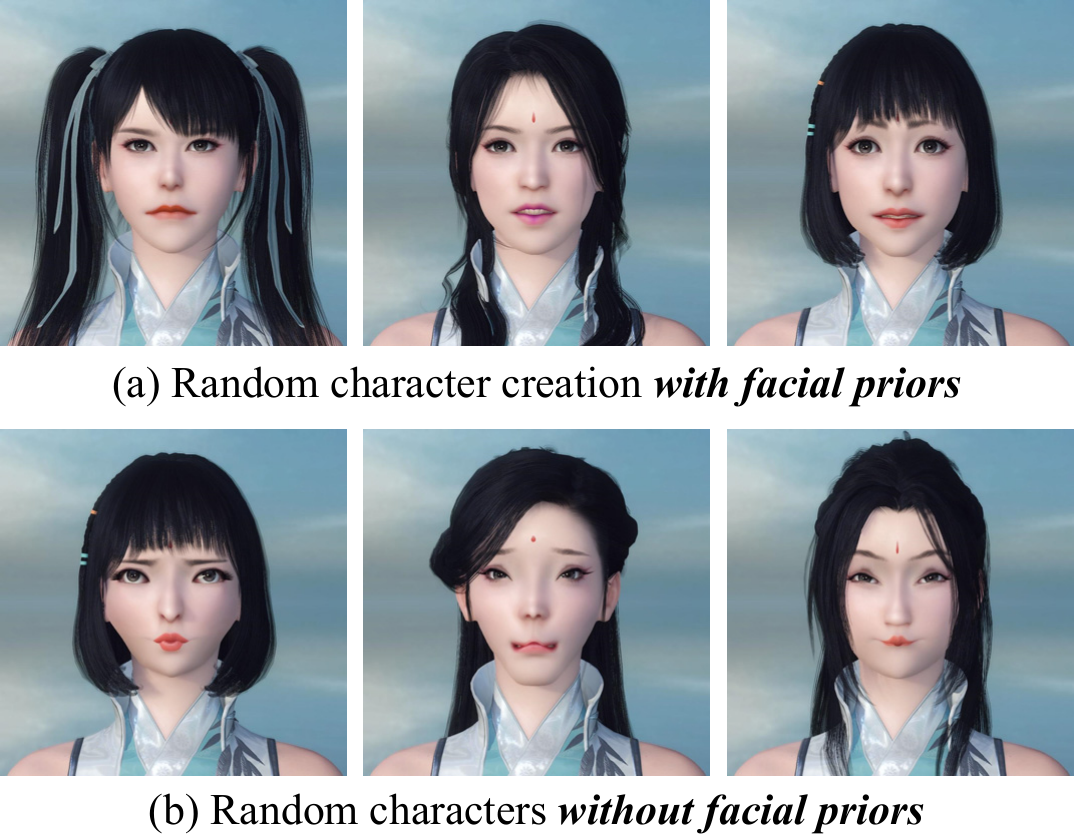}
	\caption{Randomly generated characters (a) with and (b) without integrating facial priors. The facial priors can be considered as an additional constraint on the predicted parameters and prove to help generate more stable and meaningful characters.}
	\label{fig:facial_priors}
\end{figure}

\subsection{Integrating facial priors}

To further improve the robustness of our framework, instead of predicting facial parameters directly in the parameter space, we perform a whitening transform on the facial parameters and predict in their intrinsic low-dimensional subspace. This can be considered as an integration of facial priors or an additional constraint on the predicted parameters. We apply F2P \cite{Shi_2019_ICCV} on CelebA \cite{liu2015faceattributes}, a large-scale face attributes dataset with over 200k facial images and obtain a large set of facial parameters\footnote{Since F2P is sensitive to head pose, we only use the frontal faces in CelebA training set ($\sim$18k images) to learn the whitening matrix.}.

\begin{figure}[t]
	\centering
	\includegraphics[width=0.95\linewidth]{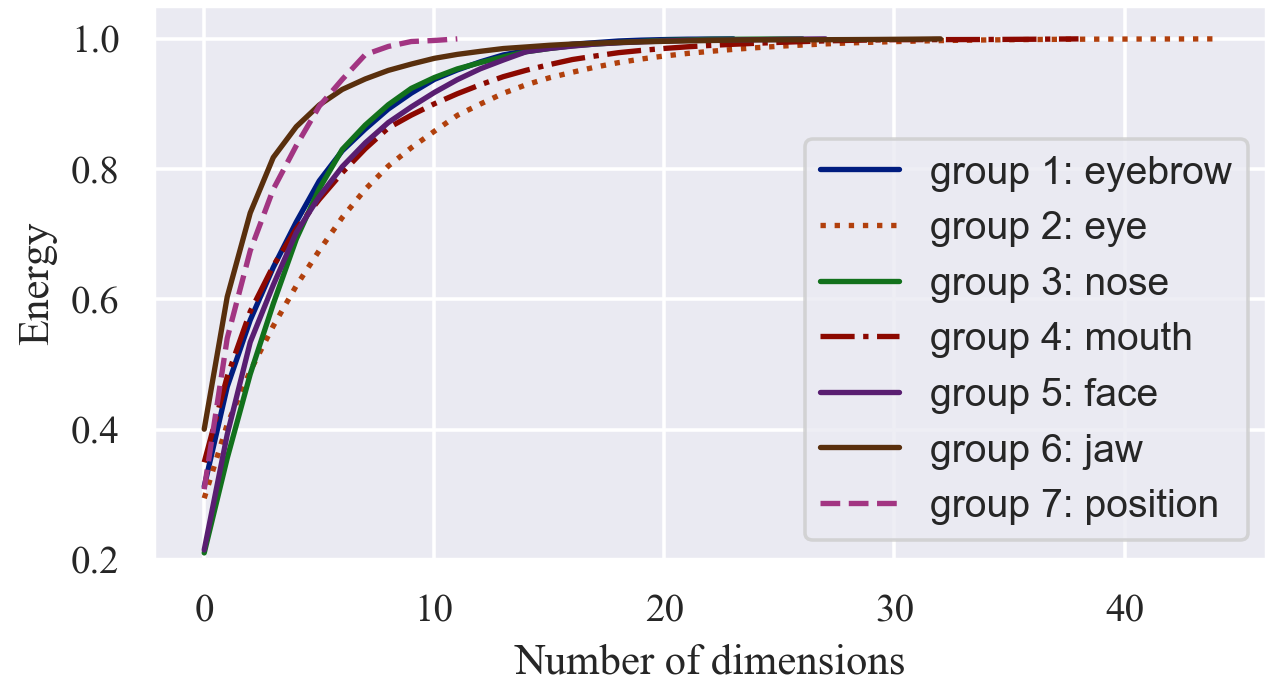}
	\caption{Reconstruction energy vs. Number of principal components in the parameter space of CelebA dataset. Regular faces are embedded in a low-dimensional subspace. We divide the facial parameters into seven groups and perform PCA dimension reduction accordingly.}
	\label{fig:engery_dims}
\end{figure}

Suppose $\bm{m}$ is the mean of facial parameters and $\bm{P}$ is the projection matrix, which can be easily obtained by performing singular value decomposition on the parameter data. Before feeding the facial parameter $\bm{x}$ into the imitator, we first project it into a subspace by multiplying with the projection matrix (whitening transformation with reduced dimensions): 
\begin{equation}
    \hat{\bm{x}}=\bm{P}^T(\bm{x}-\bm{m}).
\end{equation}
The final predicted facial parameters can be reconstructed as follows:
\begin{equation}
    \bm{x} = (\bm{P}\bm{P}^T)^{-1}\bm{P}\hat{\bm{x}} + \bm{m}.
\end{equation}
Fig.~\ref{fig:facial_priors} gives an illustration of the importance of facial prior and how it affects the rendering results. Fig.~\ref{fig:engery_dims} plots the reconstruction energy with different number of subspace dimension. With the above operation, the dimension of original facial parameters can be significantly reduced and the model becomes much easier to train.

\subsection{Facial Similarity Measurement}

After we obtain a well-trained imitator $G$ and a facial parameter translator $T$, the parameter generation essentially becomes a face similarity measurement problem. The key to our self-supervised learning framework is to \textit{enforce} the facial representation of the rendered image $I^\prime$ to be similar to that of its input face photo $I$:
\begin{equation}
    I^\prime =G(T(F_{recg}(I))) \approx I.
\end{equation}

As the input face photo and the rendered game character belong to different image domains, we introduce additional two networks, a pre-trained face recognition network $F_{recg}$ and a facial segmentation network $F_{seg}$, to learn effective facial similarity measurement in terms of both identity and local details. 

\textbf{Facial identity loss}. We use a popular face recognition network named LightCNN-29v2 \cite{wu2018light} as our face recognition network. We use its backbone to extract 256-d face embeddings as a representation of facial identity. We define the identity loss of two faces as the cosine distance on their embeddings:
\begin{equation}\label{eq:idt_loss}
\mathcal{L}_{idt} = 1 - \bm{e}_1^T\bm{e}_2/\sqrt{\|\bm{e}_1\|_2^2 \|\bm{e}_2\|_2^2},
\end{equation}
where $\bm{e}_i$, $i=1,2$ are the face embeddings of an input face photo $I$ and a rendered face image $I^\prime$:
\begin{equation}
        \bm{e}_1 = F_{recg}(I), \ \ \bm{e}_2 = F_{recg}(I^\prime).
\end{equation}

\textbf{Facial content loss}. In addition to the facial identity loss, we also define a content loss by computing pixel-wise distance based on the facial representations extracted from a pre-trained face semantic segmentation model. The motivation behind is intuitive, i.e., if two faces share the same displacement of different facial components, e.g. the contour of the face, the eyes, and nose, they will share a high similarity regardless of the image domains. We build our facial segmentation model $F_{seg}$ based on Resnet-50 \cite{he2016deep}. We remove its fully connected layers and increase its output resolution from 1/32 to 1/8. We train this model on Helen face semantic segmentation dataset \cite{le2012interactive} and then freeze its weights as a fixed feature extractor. We define the facial content loss on the pixel-wise $l_1$ distance between the semantic representations of two faces:
\begin{equation}\label{eq:ctt_loss}
\mathcal{L}_{ctt} = E_k\{\norm{\bm{f}_1 - \bm{f}_2}_1 \},
\end{equation}
where $k$ is the pixel location of the feature map. $\bm{f}_i$, $i=1,2$ are the facial semantic features of the image $I$ and $I^\prime$:
\begin{equation}
    \bm{f}_1 = F_{seg}(I), \ \ \bm{f}_2 = F_{seg}(I^\prime).
\end{equation}

\textbf{Loopback loss}. Inspired by the unsupervised 3D face reconstruction method proposed by Genova \textit{et al.} \cite{genova2018unsupervised}, we also introduce a ``loopback loss'' to further improve the robustness of our parameter output. After we obtain the face embeddings of the rendered image $F_{recg}(I^\prime)$, we further feed it into our translator $T$ to produce a set of new parameters $x^\prime=T(F_{recg}(I^\prime))$ and force the generated facial parameters before and after the loop unchanged. The loopback loss can be written as follows:
\begin{equation}\label{eq:loop_loss}
\mathcal{L}_{loop} = \norm{\bm{x} - T(F_{recg}(I^\prime))}_1.
\end{equation}

The final loss function in our model can be written as the summary of the above three losses:
\begin{equation}
\mathcal{L}(G, T, F_{recg}, F_{seg}) = \lambda_1 \mathcal{L}_{idt} + \lambda_2 \mathcal{L}_{ctt} + \lambda_3 \mathcal{L}_{loop},
\end{equation}
where $\lambda_i>0$, $i=1,2,3$ control the balance between them. Since our imitator $G$, face recognition network $F_{recg}$ and face segmentation networks $F_{seg}$ are all pre-trained models. All we need to train is the translator $T$. We aim to solving the following optimization problem:
\begin{equation}\label{eq:T_loss}
    T^{\star} = \text{argmin}_{T} \ \ \mathcal{L}(G, T, F_{recg}, F_{seg}).
\end{equation}

A complete implementation pipeline of our method is summarized as follows:\par
\begin{itemize}
\item \textbf{Training (I).} Train the imitator $G$ by minimizing Eq (\ref{eq:G_loss}). Train the face recognition network $F_{recg}$ and the face segmentation network $F_{seg}$.
\item \textbf{Training (II).} Fix $G$, $F_{recg}$, $F_{seg}$, and train the translator $T$ according to Eq (\ref{eq:T_loss}).
\item \textbf{Inference.} Given an input photo $I$, predict the facial parameters based on Eq (\ref{eq:T_inference}): $\bm{x}=T(F_{recg}(I))$.
\end{itemize}\par

\begin{table*}[t]
\centering
\caption{Performance comparison. We compare our method with two previous methods on seven face verification datasets. Higher scores indicate better. (Face embeddings are normalized by PCA as 3DMM-CNN for a fair comparison)}
\begin{tabular}{L{3.0cm}|ccccccc|C{1.5cm}}
\toprule
\multirow{2}{*}{\textbf{Method}} & \multicolumn{7}{c|}{\textbf{Datasets}}& \multirow{2}{*}{\textbf{Speed$^*$}}\\ 
 & LFW & CFP\_FF & CFP\_FP & AgeDB & CALFW & CPLFW & Vggface2\_FP & \\
\midrule 
3DMM CNN & 0.9235 & - & - & - & - & - & - & $\sim10^2$Hz \\
F2P & 0.6977 & 0.7060 & 0.5800 & 0.6013 & 0.6547 & 0.6042 & 0.6104 & $\sim1$Hz \\
Ours & {\bf 0.9402} & {\bf 0.9450} & {\bf 0.8236} & {\bf 0.8408} & {\bf 0.8463} & {\bf 0.7652} & {\bf 0.8190} & $\bf{\sim10^3}$\textbf{Hz} \\
\midrule
LightCNN-29v2$^{**}$ & 0.9958 & 0.9940 & 0.9494 & 0.9597 & 0.9433 & 0.8857 & 0.9374 & $\sim10^3$Hz\\
\bottomrule
\multicolumn{9}{l}{* Inference time under GTX 1080Ti. The time cost for model loading and face alignment are not considered.}\\
\multicolumn{9}{l}{** Here we use the performance of LightCNN-29v2 on input photos as a reference. (upper-bound accuracy)}\\
\end{tabular}%
\label{tab:comparison}
\end{table*}%

\begin{table*}[t]
\centering
\caption{Ablations analysis on different technical components of our method. The integration of each technique yields consistent improvements in face verification accuracy.}

\begin{tabular}{C{1.0cm}C{1.0cm}C{1.0cm}C{1.0cm}|ccccccc}
\toprule
\multicolumn{4}{c|}{\textbf{Ablations}} & \multicolumn{7}{c}{\textbf{Datasets}}\\
$\mathcal{L}_{ctt}$ & $\mathcal{L}_{idt}$ & $\mathcal{L}_{loop}$ & ResAtt & LFW & CFP\_FF & CFP\_FP & AgeDB & CALFW & CPLFW & Vggface2\_FP \\
\midrule 
$\checkmark$ & $\times$ & $\times$ & $\checkmark$  &  0.7880 & 0.7930 & 0.6666 & 0.6868 & 0.6792 & 0.6252 & 0.6696  \\
$\checkmark$ & $\checkmark$ & $\times$ & $\checkmark$ & 0.8843 & 0.8777 & 0.7507 & 0.7917 & 0.7675 & 0.7032 & 0.7432  \\
$\checkmark$ & $\checkmark$ & $\checkmark$ & $\times$ & 0.8870 & 0.8901 & 0.7626 & 0.7875 & 0.7725 & 0.7042 & 0.7618  \\
$\checkmark$ & $\checkmark$ & $\checkmark$ & $\checkmark$ & {\bf 0.9243} & {\bf 0.9200} & {\bf 0.7896} & {\bf 0.8152} & {\bf 0.8130} & {\bf 0.7400} & {\bf 0.7854} \\
\midrule 
\multicolumn{4}{c|}{LightCNN-29v2$^*$} & 0.9948 & 0.9939 & 0.9476 & 0.9537 & 0.9438 & 0.8872 & 0.9326 \\
\bottomrule
\multicolumn{11}{l}{* Here we use the performance of LightCNN-29v2 on input photos as a reference. (upper-bound accuracy)}\\
\end{tabular}%
\label{tab:ablation}
\end{table*}%

\subsection{Training Details}

We use the SGD optimizer with momentum to train $G$ and $F_{seg}$ with the learning rate of $10^{-2}$ and $10^{-3}$, respectively. We set $\text{batch\_size}=16$, $\text{momentum}=0.9$. The learning rate decay is set to 10\% per 50 epochs and the training stops at 500 epochs. We then freeze above networks, set $\lambda_1=0.01$, $\lambda_2=1$, and $\lambda_3=1$, and use the Adam optimizer \cite{kingma2014adam} to train $T$ with the learning rate of $10^{-4}$ and max-iteration of 20 epochs.\par

We use the CelebA dataset \cite{liu2015faceattributes} to train our translator $T$. We set $\lambda_2=0$ every 4 training steps. When the $\lambda_2$ is set to 0, we update the $T$ by sampling from the full CelebA training set, while when the $\lambda_2$ is set $>0$, we update the $T$ by sampling from a subset of the CelebA training set, which only contains high-quality frontal faces. In this way, our translator not only can be well-trained on frontal images with the content loss, but also can generalize to the rest images with the help of the identity loss, which is pose-invariant.\par

The face alignment is performed by using dlib library \cite{dlib09} to align the input photo before it is fed into our framework.\par

\begin{figure*}[t]
	\centering
	\includegraphics[width=0.88\linewidth]{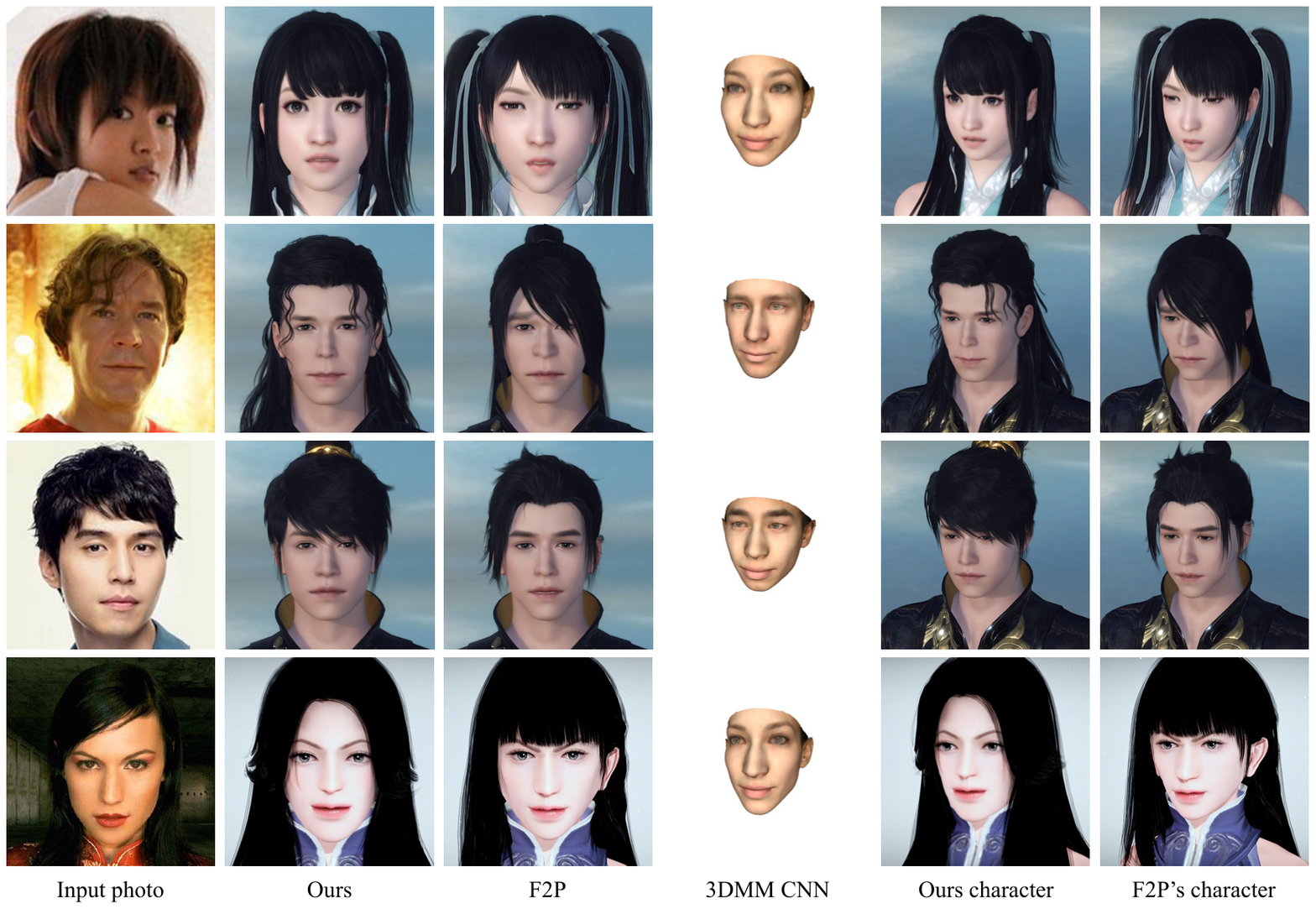}
	\caption{A comparison with other methods: 3DMM-CNN and F2P. The first three rows show results on the PC game ``Justice'' and the last row shows the results on the mobile game ``Revelation''. Our method achieves better creation results.} 
	\label{fig:comparison}
\end{figure*}

\section{Experiments}

We test our method on two games with East Asian faces, ``Justice'' and ``Revelation\footnote{https://tym.163.com/}'', where the former one is a PC game and the latter one is a mobile game on Android/IOS devices. We use seven well-known face verification datasets to quantitatively evaluate effectiveness of our method, including LFW \cite{LFW}, CFP\_FF \cite{CFP}, CFP\_FP \cite{CFP}, AgeDB \cite{Agedb}, CALFW \cite{CALFW}, CPLFW \cite{CPLFW}, and Vggface2\_FP \cite{vggface2}. \par

\subsection{Comparison with other methods}

We compare our method with some other approaches, including the F2P \cite{Shi_2019_ICCV} and 3DMM-CNN \cite{Tran_2017_CVPR} on both their accuracy and speed. Fig.~\ref{fig:comparison} shows a comparison of generated characters of different methods. Our method creates game characters with high similarity in terms of both global appearance and local details while the 3DMM method can only generate masks with similar facial outlines. 

To quantitatively evaluate the generated results, we follow the 3DMM-CNN \cite{Tran_2017_CVPR}, where the face verification accuracy is used as the evaluation metric. To do this, we first generate the facial parameters for every input face, then we use the face verification benchmark toolkit ``face.evoLVe'' \cite{zhao2019look,zhao2019multi,zhao2018towards} to compute the verification accuracy based on the generated parameters. Clearly, if the two photos belong to the same person, they should have similar facial parameters. Table~\ref{tab:comparison} shows the verification scores of different methods. A higher score suggests a better performance. The performance of 3DMM-CNN is reported by Tran \textit{et al.}'s paper \cite{Tran_2017_CVPR}. \par

\begin{table}[t]
\centering
\caption{Subjective evaluation.}
\begin{tabular}{l|c|c|c}
\toprule
Method & Selection Ratio & LightCNN & evoLVE \\
\midrule 
3DMMCNN & 17.4\% $\pm$ 1.2\% & 0.13 & 0.06  \\
F2P & 33.9\% $\pm$ 1.2\% & 0.30  & 0.18 \\
Ours & {\bf 48.7\% $\pm$ 1.1\%} & {\bf 0.34} & {\bf 0.22} \\
\bottomrule
\end{tabular}%
\label{tab:subjective}
\end{table}%

We further follow the subjective evaluation in F2P \cite{Shi_2019_ICCV} by inviting 15 volunteers to rank the results generated by three methods, and finally calculate the Selection Ratio for each method. Due to the large number of images, we only randomly collect 50 images from CelebA test set for evaluation. Table~\ref{tab:subjective} shows the subjective evaluation results and cosine similarities measured by face recognition networks (LightCNN-29v2 and face.evoLVE-IR50 \cite{zhao2019look}). Our method achieves consistent improvements over previous methods.\par

\subsection{Ablation Studies}

The ablation experiments are conducted to verify the importance of each component in our framework, including:

1) the facial identity loss $\mathcal{L}_{idt}$, (Eq. \ref{eq:idt_loss});

2) the facial content loss $\mathcal{L}_{ctt}$, (Eq. \ref{eq:ctt_loss});

3) the loopback loss $\mathcal{L}_{loop}$, (Eq. \ref{eq:loop_loss});

4) the residual attention block (ResAtt) in our facial parameter translator, (Fig.~\ref{fig:translator}). 

\begin{figure}[t]
	\centering
	\includegraphics[width=0.9\linewidth]{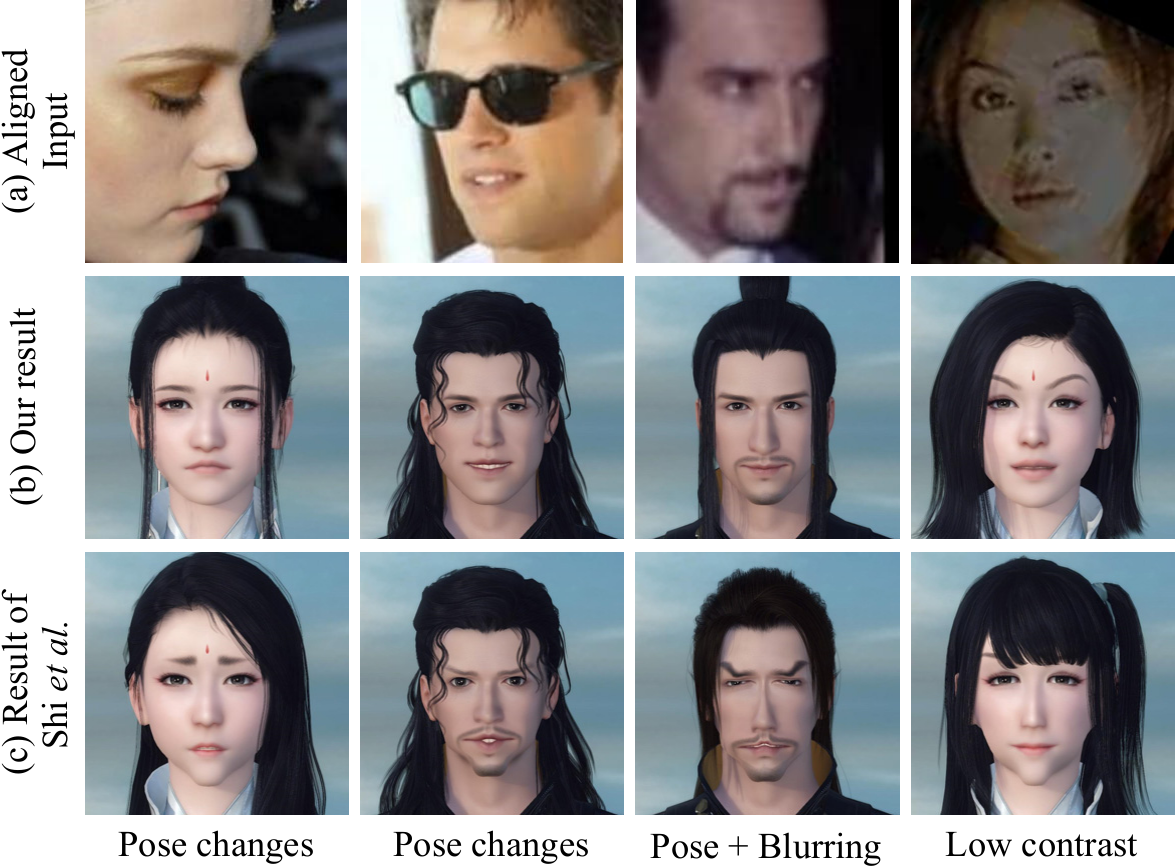}
	\caption{Comparison results on robustness.}
	\label{fig:robust_cmp}
\end{figure}

We use the same evaluation metric as we used in our comparison experiment. We first evaluate our baseline method, where we train our model only based on the facial content loss, then we gradually add other technical components. Table~\ref{tab:ablation} shows their performance on different datasets. The integration of the facial identity loss, loopback loss, and the residual attention mechanism in our translator yield consistent improvements in the face verification accuracy. Besides, the identity loss brings noticeable improvement for our method. This is because the face recognition network $F_{recg}$ and the face verification metric all focus on the facial identities. We also can observe a significant improvement by introducing the attention mechanism to our translator rather than by simply applying the vanilla multi-layer perceptron which is used in Genova's method \cite{genova2018unsupervised}.

\subsection{Experiment on robustness}
We further evaluate our method on different blurring conditions, different illuminations conditions, and the faces with different poses. Fig.\ \ref{fig:robust_cmp} shows the comparison results. Our method proves to be robust to these changes and has better robustness than the F2P, especially for the pose variants. \par

\section{Conclusion}

This paper proposes a new framework for game character auto-creation, i.e., to generate game characters of players based on their input face photos. Different from the previous methods that are designed based on neural style transfer and predicting facial parameters based on an iterative inference pipeline, we construct our method under a self-supervised learning paradigm and introduce a facial parameter translator so that the prediction can be done efficiently through a single forward propagation. Our method achieves a considerable 1000x computational speedup. Our method also proves to be more robust than previous methods, especially for the pose variants. Comparison results and ablation analysis on seven benchmark datasets suggest the effectiveness of our method.\par

{\fontsize{9.0pt}{10.0pt} \selectfont
\bibliography{FRF2P}
\bibliographystyle{aaai}	
}

\newpage
\appendix
\onecolumn
\section{Appendix}
\subsection{A description of facial parameters}
\label{facial_parameter}
Table \ref{table:facial-parameter} lists a detailed description in the PC game ``Justice'', where the ``Component'' represents the facial parts which parameters belong to, the ``Controllers'' represents user-adjustable parameters of each facial part (one controller panel contains several continuous parameters representing its shift, orientation and scale), and the ``\# parameters'' represents the parameter number of a component, of which total number is 208. Besides, there are additional 102 discrete parameters for female (22 hair styles, 36 eyebrow styles, 19 lipstick styles, and 25 lipstick colors) and 56 discrete parameters for male (23 hair styles, 26 eyebrow styles, and 7 beard styles). As for the bone-driven face model in the mobile game ``Revelation'', it has basically the same configuration of ``Controllers'' but contains less user-adjustable parameters, i.e. 174.
\begin{table}[h]
    \centering
    \begin{tabular}{l|L{10cm}|c|r}
        \toprule
        Component & Controllers &  \# parameters & Sum \\
        \midrule
        Eyebrow & eyebrow-head, eyebrow-body, eyebrow-tail & 24 & \multirow{5}{*}{208}\\
        Eye & whole, outside upper eyelid, inside upper eyelid, lower eyelid, inner corner, outer corner &  51 & \\
        Nose & whole, bridge, wing, tip, bottom &  30 & \\
        Mouth & whole, middle upper lip, outer upper lip, middle lower lip, outer lower lip, corner & 42 & \\
        Face & forehead, glabellum, cheekbone, risorius, cheek, jaw, lower jaw, mandibular corner, outer jaw & 61 & \\
        \bottomrule
    \end{tabular}
    \caption{A detailed interpretation of facial parameters (continuous part).}
    \label{table:facial-parameter}
\end{table}
\subsection{Configurations of our translator}
In our translator, we adopt three residual attention blocks and two fully connected layers. As shown in Sec. \ref{facial_parameter}, the translator need predict two kinds of parameters, i.e. continuous and discrete parameters, thus we modify the output fully connected layer into two heads: \par
1. Continuous head. This head firstly predicts a vector in an orthogonal space, then this vector is projected into the original parameter space by Eq (6) (main paper) as continuous parameters. \par
2. Discrete head. This head predicts a vector containing several groups, each group corresponds a kind of makeup, e.g. hairstyle, eyebrow style, and etc. Then we use the \emph{softmax} function to normalize this vector group-by-group as the discrete parameters.\par
In the end, we concatenate the above outputs of two heads, i.e. continuous and discrete parameters, as facial parameters and feed them into our imitator for rendering. The detailed configuration is shown in Table~\ref{table:networks-T}, ``Configuration'' lists detailed parameter settings in each layer, ``Linear(in\_dim, out\_dim)'' represents the input and output channel number of fully connected layers, ``Res-Att(in\_dim, out\_dim)'' represents the input and output channel number of residual attention blocks, ``cn'' and ``dn'' represent the dimensions of continuous and discrete parameters respectively. Inspired by ResNet \cite{he2016deep} and SENet \cite{hu2018squeeze}, the four fully connected layers in residual attention blocks (as shown in Fig. 3, main paper) are orderly set to ``Linear(512,1024)'', ``Linear(1024,512)'', ``Linear(512,16)'' and ``Linear(16,512)''.\par

\begin{table}[h]
\centering
\renewcommand\arraystretch{1.2}
\begin{tabular}{c|c|c|c}
     \toprule  
     \hline 
      \textbf{Layer} & \multicolumn{2}{c|}{\textbf{Configuration}} & \textbf{Output Size}\\
     \hline
     \hline
      FC\_1 & \multicolumn{2}{c|}{Linear(256,512)} & 512 \\
      \hline
      RA\_2 & \multicolumn{2}{c|}{Res-Att(512,512)}  & 512 \\
      \hline
      RA\_3 & \multicolumn{2}{c|}{Res-Att(512,512)} & 512 \\
      \hline
      RA\_4 & \multicolumn{2}{c|}{Res-Att(512,512)} & 512 \\
      \hline
      FC\_5 & Linear(512, cn) + BatchNorm & Linear(512, dn) &  cn + dn\\
    \hline
     \bottomrule
 \end{tabular}
     \caption{A detailed configuration of our Translator $T$.}
     \label{table:networks-T}
 \end{table}

\newpage
\subsection{More comparison results -- Female character in the PC game ``Justice''}

\begin{figure}[ht]
    \centering{\includegraphics[width=0.9\linewidth]{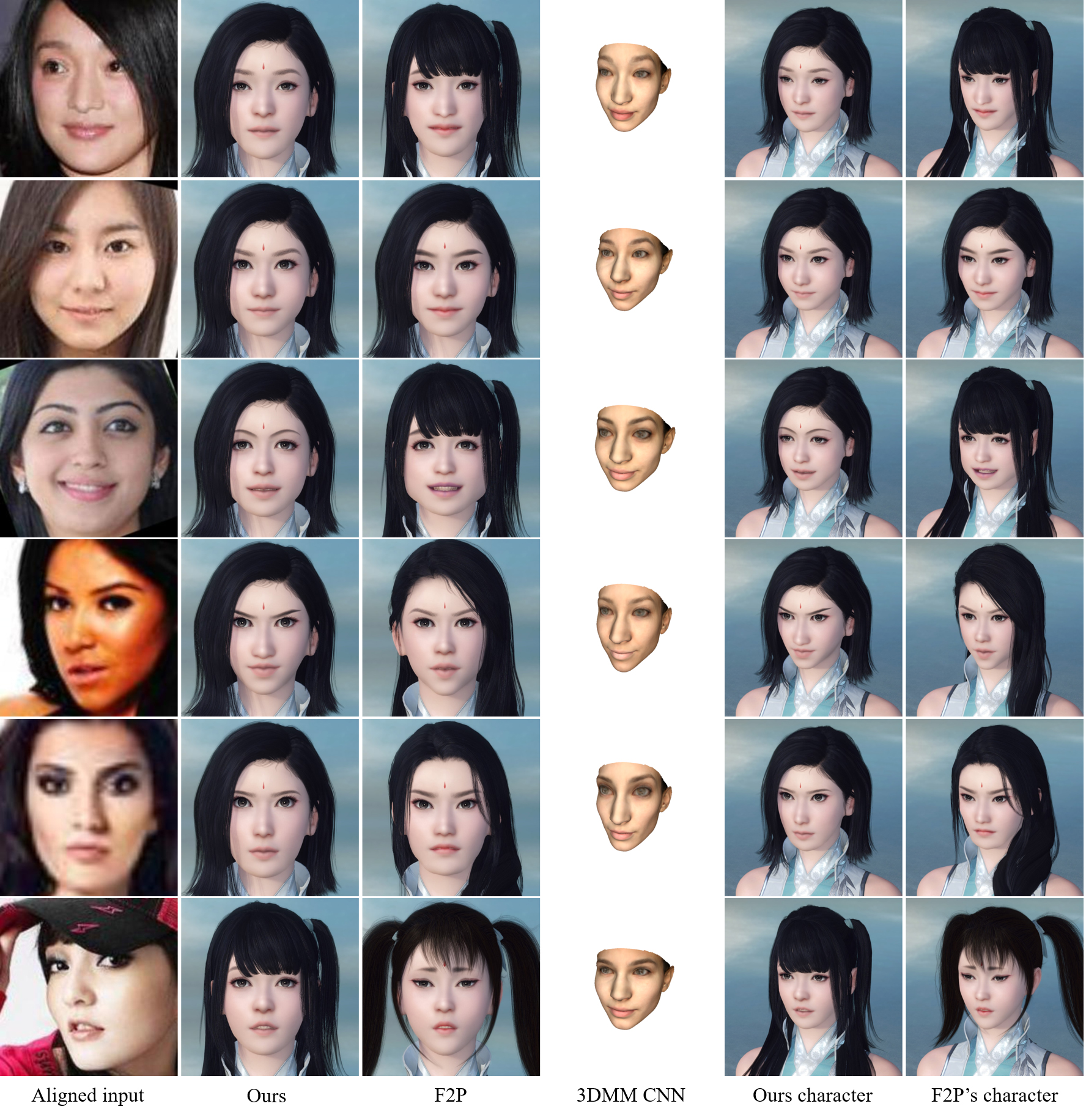}} \\
    \caption{More comparison results with other methods: 3DMM-CNN \cite{Tran_2017_CVPR} and F2P \cite{Shi_2019_ICCV}.}
    \label{fig:comparison_pc_female}
\end{figure}

\newpage
\subsection{More comparison results -- Male character in the PC game ``Justice''}

\begin{figure}[ht]
    \centering{\includegraphics[width=0.9\linewidth]{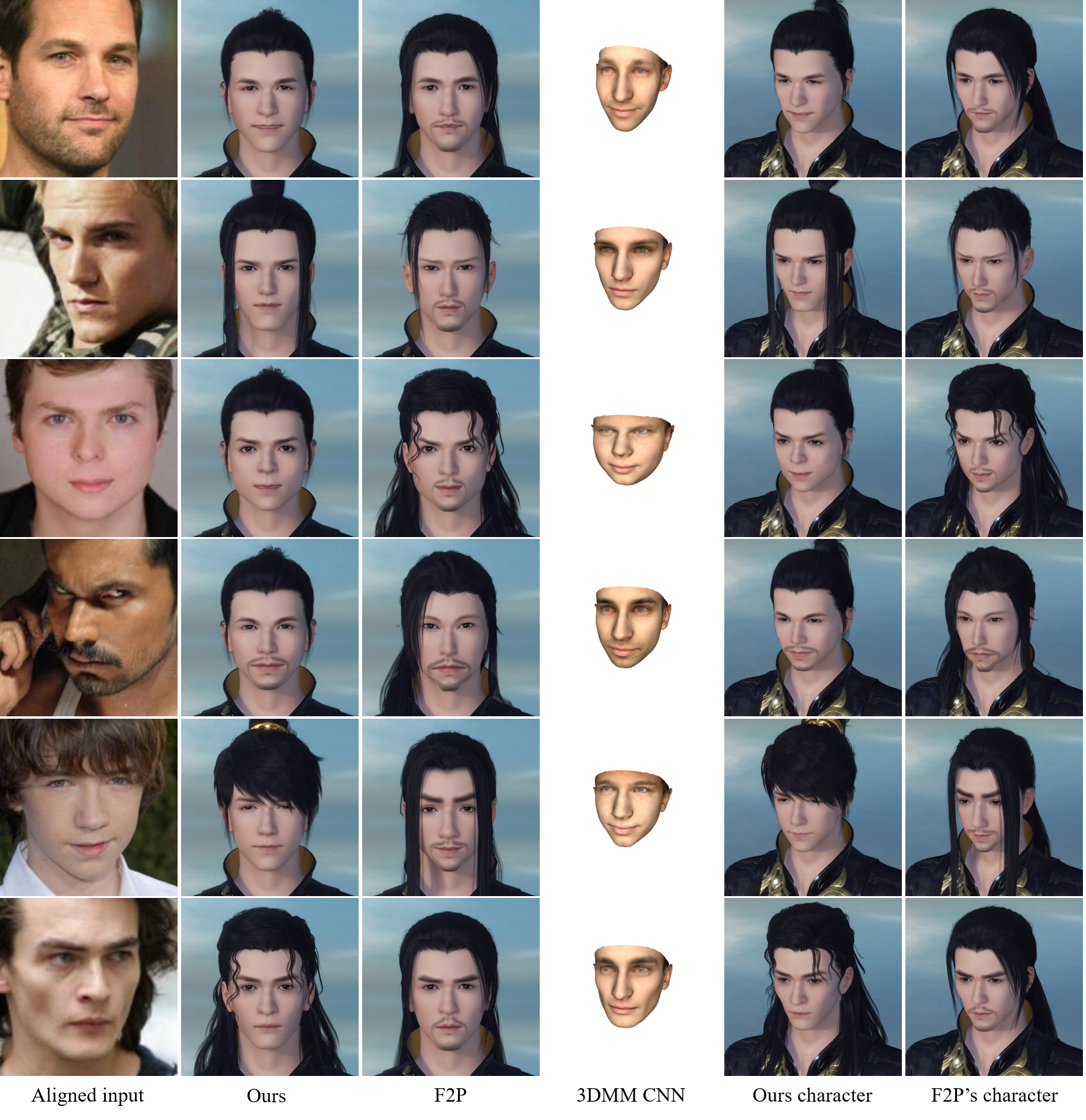}} \\
    \caption{More comparison results with other methods: 3DMM-CNN \cite{Tran_2017_CVPR} and F2P \cite{Shi_2019_ICCV}.}
    \label{fig:comparison_pc_male}
\end{figure}

\newpage
\subsection{More comparison results -- Female character in the mobile game ``Revelation''}

\begin{figure}[ht]
    \centering{\includegraphics[width=0.9\linewidth]{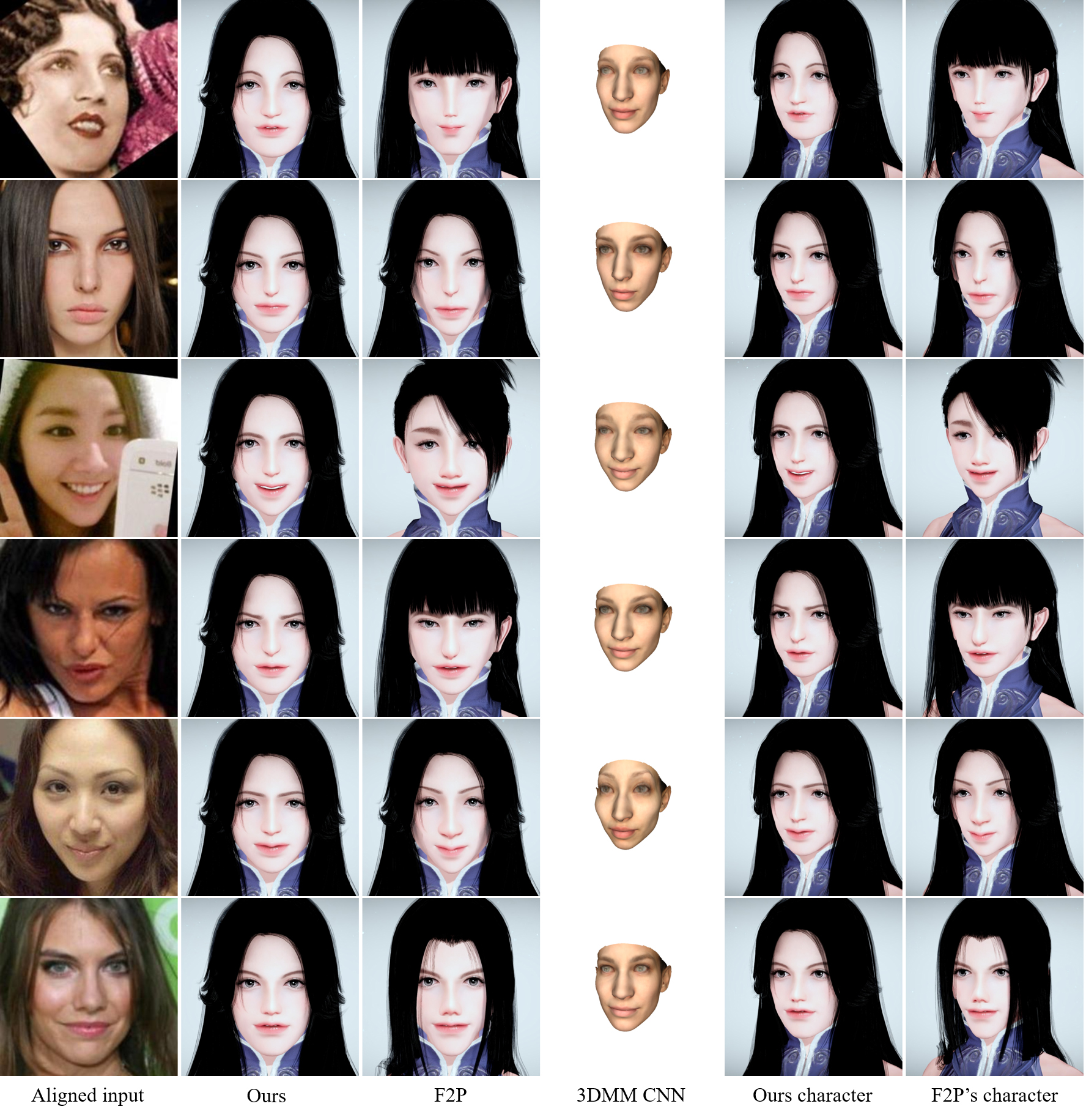}} \\
    \caption{More comparison results with other methods: 3DMM-CNN \cite{Tran_2017_CVPR} and F2P \cite{Shi_2019_ICCV}.}
    \label{fig:comparison_mobile_female}
\end{figure}

\end{document}